\documentclass[10pt,twocolumn,letterpaper]{article}

\usepackage{cvpr}  
\definecolor{cvprblue}{rgb}{0.21,0.49,0.74}
\usepackage[pagebackref,breaklinks,colorlinks,allcolors=cvprblue]{hyperref}

\def\paperID{6240} 
\def\confName{CVPR}
\def\confYear{2025}

\title{InstanceGaussian: Appearance-Semantic Joint Gaussian Representation \\for 3D Instance-Level Perception}

\usepackage{times}
\usepackage{epsfig}
\usepackage{graphicx}
\usepackage{amsmath}
\usepackage{amssymb}
\usepackage{dsfont}

\usepackage{algorithm}
\usepackage{algpseudocode}
\usepackage{tikz}

\usepackage{amsmath}
\usepackage{amsfonts}

\usepackage{booktabs}
\usepackage{pifont}
\usepackage{multirow}

\begin{document}

\vspace{-5pt}
\author{
{Haijie Li}\textsuperscript{\rm 1} 
~~~{Yanmin Wu}\textsuperscript{\rm 1} 
~~~{Jiarui Meng}\textsuperscript{\rm 1} 
~~~{Qiankun Gao}\textsuperscript{\rm 1}
~~~{Zhiyao Zhang}\textsuperscript{\rm 3} \\
\vspace{6pt}
~~~{Ronggang Wang}\textsuperscript{\rm 1,2}
~~~{\href{https://jianzhang.tech/}{Jian Zhang}}\textsuperscript{\rm 1,2*} \\
$^1$School of Electronic and Computer Engineering, Peking University, China \\
$^2$Guangdong Provincial Key Laboratory of Ultra High Definition Immersive Media Technology,\\ 
Shenzhen Graduate School, Peking University \\
$^3$College of Information Science and Engineering, Northeastern University, China
}

{
\small
\bibliographystyle{plain}
}
\maketitle

\begin{abstract}
3D scene understanding is vital for applications in autonomous driving, robotics, and augmented reality. However, scene understanding based on 3D Gaussian Splatting faces three key challenges: \textit{(i)} an imbalance between appearance and semantics, \textit{(ii)} inconsistencies in object boundaries, and \textit{(iii)} difficulties with top-down instance segmentation.
To address these challenges, we propose \textbf{InstanceGaussian}, a method that jointly learns appearance and semantic features while adaptively aggregating instances. Our contributions are as follows: \textit{(i)} a new Semantic-Scaffold-GS representation to improve feature representation and boundary delineation, \textit{(ii)} a progressive training strategy for enhanced stability and segmentation, and \textit{(iii)} a category-agnostic, bottom-up instance aggregation approach for better segmentation. Experimental results demonstrate that our approach achieves state-of-the-art performance in category-agnostic, open-vocabulary 3D point-level segmentation, validating the effectiveness of our proposed method. Project page: \href{https://lhj-git.github.io/InstanceGaussian/}{https://lhj-git.github.io/InstanceGaussian/}

\end{abstract}

\section{Introduction}
\renewcommand{\thefootnote}{}
\footnotetext{$*$ Corresponding author}
\footnotetext{This work is supported by Guangdong Provincial Key Laboratory of Ultra High Definition Immersive Media Technology(Grant No. 2024B1212010006)}
3D scene understanding aims to comprehensively capture a scene's content, structure, and semantics, addressing tasks like instance segmentation, object detection, and open-vocabulary queries. This is crucial for applications in autonomous driving, robotics, and augmented reality, where detailed scene comprehension can significantly improve performance and safety. Traditional 3D representations, such as voxels, point clouds, and meshes, provide structured models but often involve trade-offs between spatial resolution and computational efficiency. 3D Gaussian Splatting (3DGS)~\cite{kerbl20233d} offers a novel approach that combines the explicitness of traditional representations with neural adaptability, enabling efficient and detailed scene modeling. 3DGS is particularly promising for advancing 3D scene understanding by effectively balancing quality and efficiency in capturing complex scene geometry.

In the realm of 3D scene understanding utilizing 3DGS, three critical challenges emerge. 
\textit{1)} The first is an inherent \textbf{imbalance} in the representation of appearance versus semantics. Capturing the fine-grained texture details of a given object or region necessitates multiple Gaussians, each with varied appearance attributes. However, these Gaussians collectively require only a single, shared semantic attribute to accurately represent the semantics of the region. This imbalance results in a scenario where the number of Gaussians may be sufficient for appearance representation yet redundant for semantic expression. 
\textit{2)} The second challenge is an \textbf{inconsistency} between appearance and semantics. Without semantic constraints, \textit{i.e.}, in pure appearance reconstruction, a single Gaussian can represent different objects or regions. At object boundaries, for instance, a single Gaussian may simultaneously represent both the object's foreground and background. Recent approaches, such as language-embedded Gaussian Splatting techniques~\cite{qin2024langsplat,NEURIPS2024_21f7b745}, employ a decoupled learning strategy that overlooks the interdependence between color and semantics, leading to inconsistencies between appearance and semantics and posing significant challenges for both 3D point segmentation and 2D image segmentation. 
\textit{3)} Thirdly, previous approaches are predominantly designed in a top-down manner, often relying on prior category information. For instance, GaussianGrouping~\cite{gaussian_grouping} defines the number of instances based on 2D tracking results, FastLGS~\cite{ji2025fastlgs} uses the count of matches in cross-view settings to determine the number of objects, and OpenGaussian~\cite{NEURIPS2024_21f7b745} relies on a pre-defined number of codebook entries. Such methods are prone to issues arising from uneven category distributions, often resulting in over-segmentation or under-segmentation when handling fine-grained instances in complex scenes.

To address these challenges, we propose \textbf{InstanceGaussian}, a method capable of jointly learning object appearance and instance features while adaptively aggregating objects. Our contributions are summarized as follows:

\begin{enumerate}
    \item \textbf{Semantic-Scaffold-GS Representation}: We introduce a novel representation method that balances appearance and semantics. By more flexibly allocating the semantic and appearance attributes of Gaussians, this method enables the learning of more accurate object geometric boundaries and yields improved feature representations.
    \item \textbf{Progressive Appearance-Semantic Joint Training Strategy}: We propose a progressive training strategy that optimizes the joint representation of appearance and semantics* in a step-by-step manner. This ensures their consistency throughout the training process, thereby enhancing the model's stability and improving the accuracy of downstream tasks such as 3D point segmentation and 2D image segmentation.
    \item \textbf{Bottom-Up, Category-Agnostic Instance Aggregation}: To further enhance the capability of 3D instance segmentation, we introduce a bottom-up, category-agnostic instance aggregation method. This method employs a clustering algorithm based on farthest point sampling and connected component analysis, effectively avoiding over-segmentation or under-segmentation.
\end{enumerate}
\footnotetext{$*$ The term ``semantic'' is used here to distinguish it from ``appearance''. In subsequent sections and code implementations, this concept will be explicitly referred to as ``instance features''.}
\section{Related Work}

\subsection{3D Representation}
3D representation involves the digital description and modeling of objects and scenes in three-dimensional space using techniques such as voxels, point clouds, meshes, signed distance functions, and neural representations. Neural representations have made substantial progress with techniques like NeRF~\cite{mildenhall2021nerf}, which has notably enhanced the quality of novel view synthesis through learning-based optimization. Various methods~\cite{barron2021mip, barron2023zip, mip360,jiang2023alignerf} have been proposed to further enhance NeRF's expressive capabilities, but they often face challenges with slow training and rendering speeds. To address these issues, alternative approaches focusing on explicit representations have been developed, which have been effective in reducing the computational burden and decreasing training and inference times. These methods include voxels~\cite{sun2022direct, fridovich2022plenoxels, reiser2023merf}, hash grids~\cite{muller2022instant} and point clouds~\cite{yifan2019differentiable, aliev2020neural, ruckert2022adop}.

3D Gaussian Splatting (3DGS)~\cite{kerbl20233d} is an innovative representation technique for modeling and rendering 3D scenes using
Gaussian distributions. In recent advancements~\cite{scaffoldgs, meng2024mirror, yu2024mip, huang20242d, yan2024gs,hicom2024}, Scaffold-GS~\cite{scaffoldgs} forms a hierarchical and region-aware scene representation that respects the underlying geometric structure of the scene. This strategy exploits scene structure to guide and constrain the distribution of 3D Gaussians, allowing them to locally adapt to varying viewing angles and distances. This innovation allows for enhanced adaptation to various viewpoints and distances while preserving the scene's underlying structure, showcasing the ongoing advancements in 3DGS and its role in shaping the future of neural rendering.

\subsection{3D Perception via 2D-3D Correlation}

Recent advancements in 3D perception have seen the integration of 2D Vision-Language Models (VLMs) with 3D point cloud processing, leading to significant progress in the field~\cite{huang2023clip2point, xue2024ulip, zhu2023pointclip,wu2022eda,wu2025language}. These methods primarily focus on aligning features and projecting 3D data into 2D to enhance zero-shot learning capabilities. Existing approaches that exploit the 2D-3D correlation can be broadly categorized into two types: one leverages feature information from 2D images, while the other utilizes segmentation masks derived from 2D images.

\textbf{Feature Distillation}.
The significant advancements in 2D scene understanding, initiated by SAM~\cite{sam} and its variants, have inspired integrating semantic features into NeRF. Various methods have been developed to incorporate semantic features from models like CLIP~\cite{clip} and DINO~\cite{dino} into NeRF, enhancing 3D segmentation, understanding, and editing tasks. LeRF~\cite{kerr2023lerf} distills features from accessible VLMs such as CLIP into a 3D scene represented by NeRF. Additionally, ~\cite{liu2023weakly} presents a 3D open-vocabulary segmentation pipeline using NeRF. Within the 3DGS framework, LEGaussians~\cite{shi2024language} introduces uncertainty and semantic feature attributes to each Gaussian, rendering a semantic map with corresponding uncertainties. This map is compared with quantized CLIP and DINO dense features extracted from the ground truth image. LangSplat~\cite{qin2024langsplat} employs a scene-wise language autoencoder to learn language features in the scene-specific latent space, effectively discerning clear boundaries between objects in rendered feature images. Feature3DGS~\cite{zhou2024feature} distills SAM's encoder features into 3D Gaussians and employs SAM's decoder to interpret 2D rendered feature maps for segmentation. 
However, feature distillation-based approaches face challenges such as high training costs, significant storage requirements, and slow inference times, limiting their practical application.

\begin{figure*}[t]
  \centering
  \includegraphics[width=0.9\linewidth]
    {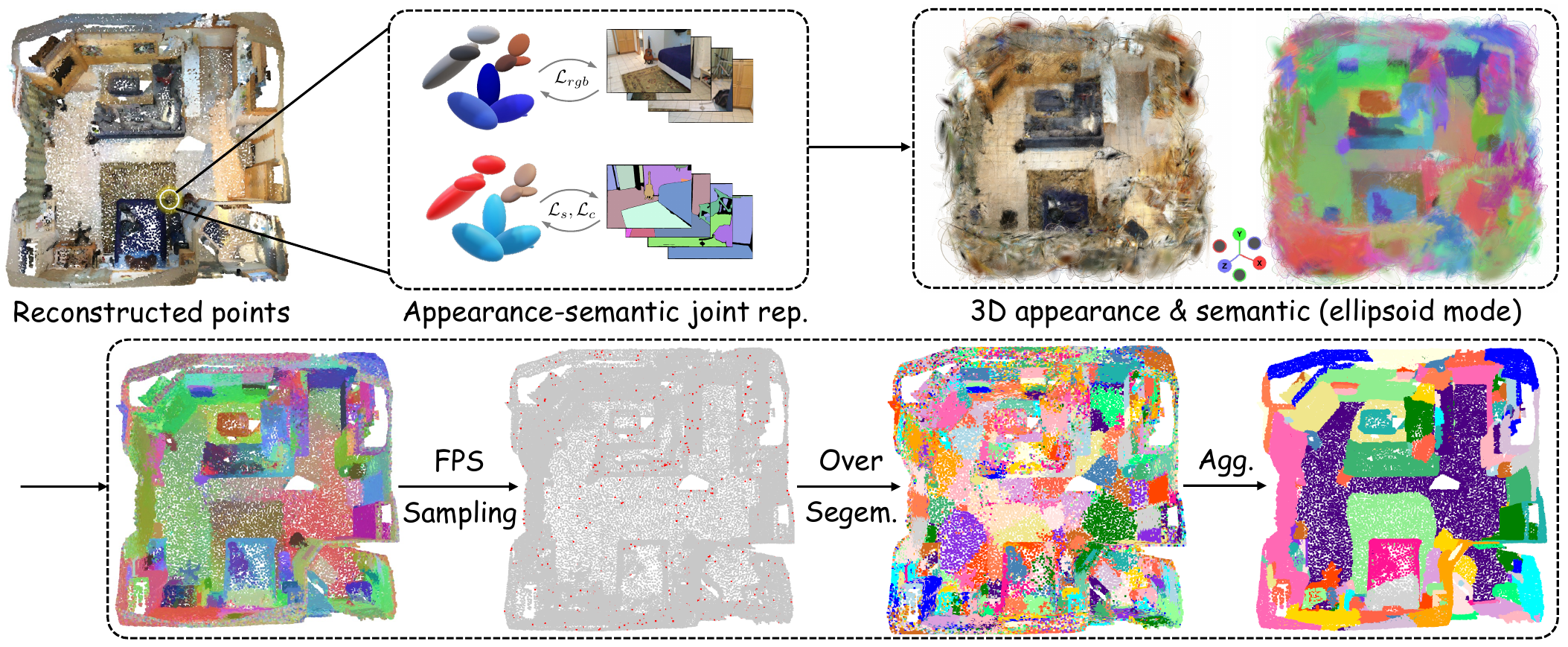}
    \vspace{-2mm}
    \caption{Top row: Appearance-semantic joint Gaussian representation avoids the imbalance and inconsistency in appearance-semantic learning. Bottom row: Bottom-up instantiation: Over-segmentation is achieved via FPS sampling and clustering, followed by instantiation through graph-connectivity-based aggregation.}
    \vspace{-4mm}
    \label{fig:main}
\end{figure*}

\textbf{2D Mask Lifting}. Recent works have explored lifting 2D segmentation masks into 3D space~\cite{yin2024sai3d, nguyen2024open3dis, yan2024maskclustering, gaussian_grouping, cen2023saga, choi2024click, NEURIPS2024_21f7b745}.
SAI3D~\cite{yin2024sai3d} and Open3DIS~\cite{nguyen2024open3dis} propose methods for merging 3D superpoints~\cite{felzenszwalb2004efficient} guided by predictions from SAM.  MaskClustering~\cite{yan2024maskclustering} introduces a mask graph clustering approach based on the view consensus rate, targeting open-vocabulary 3D instance segmentation. 
To achieve the 2D mask consistency across views, GaussianGrouping~\cite{gaussian_grouping} performs simultaneous reconstruction and segmentation of open-world 3D objects, guided by 2D mask predictions obtained from SAM and 3D spatial consistency constraints. 
SAGA ~\cite{cen2023saga} utilizes a contrastive learning approach with masks generated by SAM. It maps SAM’s features into a low-dimensional space using a trainable MLP, replicating these features to address inconsistency issues.
Click-Gaussian~\cite{choi2024click} introduces an efficient method for interactive segmentation of 3D Gaussians. It uses two-level granularity feature fields from 2D masks and incorporates Global Feature-guided Learning (GFL) to resolve cross-view inconsistencies. 
OpenGaussian~\cite{NEURIPS2024_21f7b745} employs SAM masks to train instance features with 3D consistency, proposing a two-stage codebook for feature discretization and an instance-level 3D-2D feature association method that links 3D points to 2D masks and CLIP features, demonstrating effectiveness in open vocabulary-based 3D object selection and understanding.
However, OpenGaussian uses a fixed codebook entry count and does not fully exploit the continuity of 3D space, which constrains its performance.

\section{Method}

\subsection{Preliminary}
\label{sec:pre}
\textbf{\indent 3D Gaussian Splatting ~\cite{kerbl20233d}.}
Given a set of $n$ 3D Gaussians $\mathcal{G}$, each with center $\boldsymbol{\mu}$, color $\boldsymbol{c}$, opacity $\boldsymbol{\sigma}$, rotation $\boldsymbol{R}$ and scale $\boldsymbol{S}$, they are rendered as an image $\boldsymbol{I}_{ren}$ at a specific viewpoint via differentiable rasterization $\mathcal{D}$. The reconstruction loss $\mathcal{L} _{rgb}$ is then calculated by comparing $\boldsymbol{I}_{ren}$ with the GT image $\boldsymbol{I}_{gt}$. This process can be simplified as shown below, where blue variables are optimizable parameters.

\begin{tikzpicture}
    \node (G) at (0, 0) {$\mathcal{G}(\textcolor{blue}{\boldsymbol{\mu}, \boldsymbol{c}, \boldsymbol{\sigma}, \boldsymbol{R}, \boldsymbol{S}})_n$};
    \node (I) at (2.8, 0) {$\boldsymbol{I}_{ren}$};
    \node (loss) at (4.5, 0) {$\mathcal{L} _{rgb}$};
    \node (l_gt) at (6.0, 0) {$\boldsymbol{I}_{gt}$};
    
    \draw[->] (G) -- node[above] {$\mathcal{D}$} (I);
    \draw[->] (I) -- (loss);
    \draw[<-] (loss) -- (l_gt);
\end{tikzpicture}
\vspace{-1mm}

\textbf{OpenGaussian~\cite{NEURIPS2024_21f7b745}.}
Based on 3DGS, OpenGaussian adds an instance feature $\boldsymbol{f} \in \mathbb{R}^6$ for each Gaussian, passes it through differentiable rasterization $\mathcal{D}$ to render feature map $\boldsymbol{M} \in \mathbb{R}^{H\times W\times 6}$, and computes intra-mask smoothness loss $\mathcal{L}_s$ and inter-mask contrastive loss $\mathcal{L}_c$ with SAM masks $\boldsymbol{B} \in \mathbb{R}^{H\times W\times 1}$. This process simplifies as follows, where only the feature $\boldsymbol{f}$ is optimized while other appearance attributes are frozen (shown in gray).

\begin{tikzpicture}
    \node (G) at (0, 0) {$\mathcal{G}(\textcolor{gray}{\boldsymbol{\mu}, \boldsymbol{c}, \boldsymbol{\sigma}, \boldsymbol{R}, \boldsymbol{S}}, \textcolor{blue}{\boldsymbol{f}})_n$};
    \node (I) at (2.6, 0) {$\boldsymbol{M}$};
    \node (loss) at (4.3, 0) {$\mathcal{L} _{s}, \mathcal{L} _{c}$};
    \node (l_gt) at (6.0, 0) {$\boldsymbol{B}$};
    
    \draw[->] (G) -- node[above] {$\mathcal{D}$} (I);
    \draw[->] (I) -- (loss);
    \draw[<-] (loss) -- (l_gt);
\end{tikzpicture}

\begin{equation}
    \setlength{\abovedisplayskip}{-5pt}
    \mathcal{L}_s=\sum_{i=1}^m\sum_{h=1}^H \sum_{w=1}^W \boldsymbol{B}_{i, h, w} \cdot\left\|\boldsymbol{M}_{:, h, w}-\bar{\boldsymbol{M}}_i\right\|^2,
    \label{eq:intra_mask_loss}
    \setlength{\belowdisplayskip}{-5pt}
\end{equation}

\begin{equation}
    \mathcal{L}_c=\frac{1}{m(m-1)}\sum_{i=1}^m\sum_{j=1, j\ne i}^m \frac{1}{\left\|\bar{\boldsymbol{M}}_i-\bar{\boldsymbol{M}}_j\right\|^2},
    \label{eq:inter_mask_loss}
\end{equation}
where $H, W$ represents the image size and $m$ corresponds to the number of SAM masks in the current view. The $\mathcal{L}_s$ aims to make each pixel feature $\boldsymbol{M}_{:, h, w}$ converge towards the average feature $\bar{\boldsymbol{M}}_i$ of its associated mask for intra-mask smoothness. The $\mathcal{L}_c$ seeks to maximize the separation between the average features $\bar{\boldsymbol{M}}_i, \bar{\boldsymbol{M}}_j$ of different masks.

\indent\textbf{Scaffold-GS~\cite{scaffoldgs}.}
It derives $n$ neural Gaussian points $\mathcal{G}$ from $n'$ anchor points $\mathcal{A}$ ($n=10n'$ in their implementation). The Gaussian centers $\boldsymbol{\mu}$ are obtained by shifting its anchor centers $\boldsymbol{\mu}'$ by an offset $\boldsymbol{o}$, while other appearance attributes are decoded from the appearance embedding $\boldsymbol{e}$ through MLPs. As shown below, the blue parts are learnable parameters, and their optimization process is similar to 3DGS.

\begin{tikzpicture}
    \small
    \node (A) at (0, 0) {$\mathcal{A}(\textcolor{blue}{\boldsymbol{\mu}'}, \textcolor{blue}{\boldsymbol{e}})_{n'}$};
    \node (G) at (2.5, 0) {$\mathcal{G}(\boldsymbol{\mu}, \underline{\boldsymbol{c}, \boldsymbol{\sigma}, \boldsymbol{R}, \boldsymbol{S}})_n$};
    \node (I) at (4.8, 0) {$\boldsymbol{I}_{\text{ren}}$};
    \node (loss) at (5.8, 0) {$\mathcal{L}_{\text{rgb}}$};
    \node (Igt) at (6.8, 0) {$\boldsymbol{I}_{\text{gt}}$};
    
    \draw[->] (G) -- node[above] {$\mathcal{D}$} (I);
    \draw[->] (I) -- (loss);
    \draw[<-] (loss) -- (Igt);

    \draw[->] (-0.2, 0.2) .. controls (-0.2, 0.8) and (1.8, 0.8) .. node[below, xshift=0pt, yshift=2pt] {$+~\textcolor{blue}{\boldsymbol{o}}$} (1.8, 0.2);
    
    \draw[->] (0.2, -0.2) .. controls (0.3, -0.8) and (2.8, -0.8) .. node[above, xshift=0pt, yshift=-2pt] {\textcolor{blue}{MLPs}} (2.8, -0.2);
\end{tikzpicture}

\subsection{Appearance-semantic joint Gaussian Representation}

\textbf{\indent (1) Appearance-semantic balanced representation.}
For the same object or region, the texture details may vary while the semantics remain consistent. Consequently, many Gaussians with diverse appearance attributes are required to represent the texture details of the region. These Gaussians require only a single semantic attribute to effectively capture the region's semantics, meaning the same number of Gaussians may under-represent appearance while over-representing semantics. We refer to this phenomenon as the \textbf{\textit{imbalance}} in the representation of appearance and semantics by Gaussians.
To this end, we propose a Semantic-Scaffold-GS representation to balance appearance and semantics:

\hspace*{-15pt}
\begin{tikzpicture}
    \small
    \node (A) at (0, 0) {$\mathcal{A}(\textcolor{blue}{\boldsymbol{\mu}'}, \textcolor{blue}{\boldsymbol{e}}, \textcolor{blue}{\boldsymbol{f}})_{n'}$};
    \node (G) at (2.5, 0) {$\mathcal{G}(\boldsymbol{\mu}, \underline{\boldsymbol{c}, \boldsymbol{\sigma}, \boldsymbol{R}, \boldsymbol{S}}, \boldsymbol{f})_n$};
    \node (I) at (4.8, 0) [align=center] {$\boldsymbol{I}_{\text{ren}},$ \\ $\boldsymbol{M}$};
    \node (loss) at (6, 0) [align=center] {$\mathcal{L}_{\text{rgb}},$ \\ $\mathcal{L} _{s}, \mathcal{L} _{c}$};
    \node (Igt) at (7.15, 0) [align=center] {$\boldsymbol{I}_{\text{gt}},$ \\ $\boldsymbol{B}$};
    
    \draw[->] (G) -- node[above] {$\mathcal{D}$} (I);
    \draw[->] (I) -- (loss);
    \draw[<-] (loss) -- (Igt);

    \draw[->] (-0.3, 0.2) .. controls (-0.3, 0.8) and (1.0, 0.8) .. node[below, xshift=-2pt, yshift=2pt] {$+~\textcolor{blue}{\boldsymbol{o}}$} (1.6, 0.2);

    \draw[->] (0.4, 0.2) .. controls (1.5, 0.8) and (3.5, 0.8) .. node[below, xshift=2pt, yshift=2pt] {share} (3.5, 0.2);
    
    \draw[->] (0.1, -0.2) .. controls (0.1, -0.8) and (2.5, -0.8) .. node[above, xshift=0pt, yshift=-2pt] {\textcolor{blue}{MLPs}} (2.5, -0.2);
\end{tikzpicture}
The blue parts are the optimizable parameters; see Sec.~\ref{sec:pre} for variable definitions. Similar to Scaffold-GS, we derive $n$ neural Gaussian points $\mathcal{G}$ from $n'$ anchor points $\mathcal{A}$ ($n=5n'$ in our implementation). Specifically, each group of 5 child Gaussians with different appearance attributes is decoded from the appearance embedding $\boldsymbol{e}$ of their common parent anchor point using MLPs. However, at the semantic level, these 5 child Gaussians share the same instance feature $\boldsymbol{f}$ as their parent anchor point. 
On the one hand, this representation strengthens the learning of appearance, especially in scenes represented by sparse points (\textit{e.g.}, ScanNet and SLAM-reconstructed scenes); on the other hand, sharing features among neighboring points significantly enhances the efficiency of semantic learning. The process is described in the top row of Fig.~\ref{fig:main}.

\begin{figure}[t]
  \centering
  \includegraphics[width=0.9\linewidth]
    {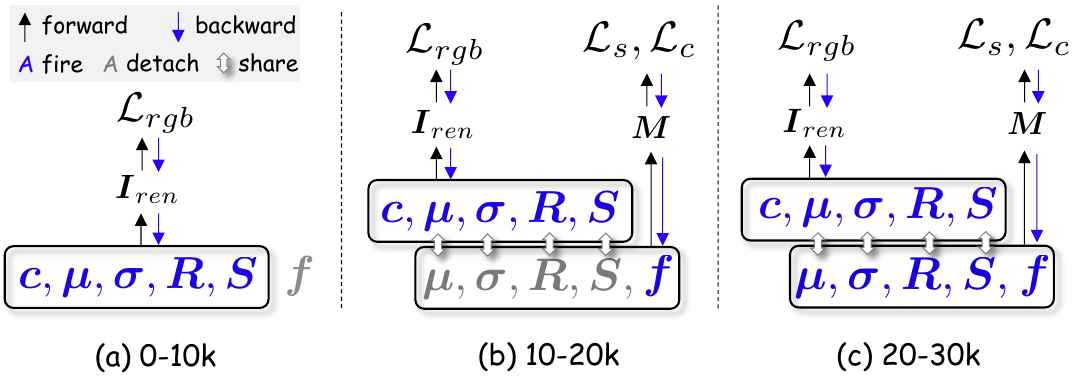}
    \vspace{-2mm}
    \caption{Progressive appearance-semantic joint training. (a) Train appearance only; (b) Independent appearance-semantic training; (c) Joint appearance-semantic training.}
    \vspace{-5mm}
    \label{fig:joint_train}
\end{figure}

\textbf{(2) Appearance-semantic consistency learning.}
Without semantic constraints, \textit{i.e.}, pure appearance reconstruction, a Gaussian can represent across objects/regions. For instance, at boundaries, a single Gaussian may simultaneously represent both the object and the background. Recent language-embedded Gaussian Splatting works~\cite{qin2024langsplat,NEURIPS2024_21f7b745} follow a paradigm of pre-training appearance reconstruction, freezing the appearance properties, and subsequently learning semantic attributes. This tends to cause \textbf{\textit{inconsistencies}} between appearance and semantics, posing challenges for 3D point segmentation and 2D image segmentation.
Therefore, we propose a progressive appearance-semantic joint training strategy to ensure their consistency as shown in Fig.~\ref{fig:joint_train}: during the 0-10k steps, only the appearance attributes are trained, then appearance and semantics are trained independently in the 10k-20k steps, and finally, appearance and semantics are jointly trained in the 20k-30k steps.
Here, we employ the loss (Eq.~(\ref{eq:intra_mask_loss})~(\ref{eq:inter_mask_loss})) of OpenGaussian to train instance features, but its inter-mask contrastive loss value is too large, causing instability of joint training. We overcome this problem through the following improvements:
\begin{equation}
    \mathcal{L}_c=\frac{1}{m(m-1)}\sum_{i=1}^m\sum_{j=1, j\ne i}^m \frac{ \mathds{1}_{\| \bar{M}_i - \bar{M}_j \|^2 < \tau}  }{\left\|\bar{\boldsymbol{M}}_i-\bar{\boldsymbol{M}}_j\right\|^2},
    \label{eq:new_loss}
\end{equation}
where $\mathds{1}_{\| \bar{M}_i - \bar{M}_j \|^2 < \tau}$ takes values of 1 or 0, aiming to truncate the loss between two objects whose dissimilarity (quantified by the L2 distance) exceeds a threshold $\tau$ (set to 0.4 in our implementation). This improvement is intuitive because there is no need to enforce that the features of two different objects are infinitely distant from each other.

\subsection{Bottom-up Category-Agnostic Instantiation}
\label{sec:method_instantiation}
We employ a bottom-up method for category-agnostic instantiation. Specifically, we first over-segment the scene into numerous sub-objects. Then, we construct a graph of these sub-objects and perform aggregation based on the graph connectivity, adaptively obtaining several independent and complete objects. The bottom row of Fig.~\ref{fig:main} describes this process.

\textbf{(1) Over-segmentation.}
Inspired by point cloud perception methods, which utilize farthest point sampling (FPS) algorithms to sample seed points for candidate proposals from dense point clouds, we achieve over-segmentation of the scene through FPS-based clustering.

We first sample $s$ seed points from all $n$ Gaussians through FPS and use them as the initial centers of $s$ clusters, where $s$ is much greater than the number of objects in the scene, with a default value of 1000. Then we start the clustering process, updating the instance feature of all Gaussians and the $s$ clusters. This process can be formulated as:
\begin{equation}
    \boldsymbol{\bar{f}}, \boldsymbol{\bar{\mu}}, \mathbf{\bar{I}} = \phi (\mathbf{X}),
    \label{eq:k-means}
\end{equation}
\begin{equation}
    \mathbf{X} = [\mathrm{PE}(\boldsymbol{\mu}) \in \mathbb{R}^{n \times d}; \boldsymbol{f} \in \mathbb{R}^{n \times 6}],
    \label{eq:concat}
\end{equation}
where $\phi(\cdot)$ denotes the k-means, $\mathbf{X} \in \mathbb{R}^{n \times (d+6)}$ is the attribute of the $n$ Gaussians participating in the clustering, which is the concatenation of the positional encoding ($\mathrm{PE}(\cdot)$) for position $\boldsymbol{\mu}$ and the instance feature $\boldsymbol{f}$. Finally, $s$ sub-object clusters are obtained, $\boldsymbol{\bar{f}} \in \mathbb{R}^{s \times 6}$ is their feature, $\boldsymbol{\bar{\mu}} \in \mathbb{R}^{s \times 3}$ is their centers, $\mathbf{\bar{I}} \in \mathbb{Z}_{[0, s)}^n $ represents the label indicating which subobject each point belongs to. During this process, only the feature $\boldsymbol{f}$ needs to be optimized, while the position $\boldsymbol{\mu}$ is frozen. 
This process is shown in lines 1-4 of Alg.~\ref{tab:algorithm}.

\renewcommand{\algorithmicrequire}{\textbf{Input:}} 
\renewcommand{\algorithmicensure}{\textbf{Output:}}
\begin{algorithm}[t]
	\caption{Bottom-up category-agnostic instantiation}
	\label{tab:algorithm}
	\begin{algorithmic}[1] 
		\Require $\boldsymbol{\mu}, \boldsymbol{f}$ - point position and feature, $s$ - the number of over-segmented sub-objects, $r$-voxel resolution, $\gamma$ - connectivity threshold.
		\Ensure well-segmented complete object $\mathbf{\hat{I}}, \mathbf{\hat{f}}$. 
		\vspace{1mm}		
		\Procedure{overSegmentation}{$\boldsymbol{\mu}, \boldsymbol{f}, s, r, \gamma$}
                \State $\boldsymbol{\bar{f}}, \boldsymbol{\bar{\mu}} \gets$ $FPS(\boldsymbol{f},\boldsymbol{\mu}, s)$ \Comment initialize $s$ centers
                \State $\mathbf{X} \gets$ concat $PE(\boldsymbol{\mu})$ and $\boldsymbol{f}$ \Comment Eq.~(\ref{eq:concat})
                \State $\mathbf{\bar{I}},\boldsymbol{\bar{f}}, \boldsymbol{\bar{\mu}} \gets$ $k$$-$$means$$(\mathbf{X}, s)$ \Comment Eq.~(\ref{eq:k-means}) \\
            \Return \Call{subObjectAgg}{$\boldsymbol{\mu}, \mathbf{\bar{I}}, \boldsymbol{\bar{f}}, s, r, \gamma$}
            \EndProcedure
            \vspace{1mm}
            
            \Procedure{subObjectAgg}{$\boldsymbol{\mu}, \mathbf{\bar{I}}, \boldsymbol{\bar{f}}, s, r, \gamma$}
                \State $\mathbf{G} \in \mathbb{R}^{s \times s} \gets$ \text{initialize the graph with zeros}
                \State $\textbf{Vox} \gets$ $Voxelization(\boldsymbol{\mu},\mathbf{\bar{I}},r)$ \Comment sub-obj. voxels
                \For{$i$ in $[0, s)$}
                    \For{$j$ in $[0, s)$}
                        \State $dis(i,j)$ $\gets$ $L2distance(\boldsymbol{\bar{f}}_i, \boldsymbol{\bar{f}}_j)$
                        \State $\mathds{1}(i,j)$ $\gets$ $near(\textbf{Vox}_i, \textbf{Vox}_j)$ 
                        \State $\mathbf{G} _{ij}$ $\gets$ $dis(i,j)$ $\cdot$ $\mathds{1}(i,j)$ \Comment Eq.~(\ref{eq:graph})
                    \EndFor
                \EndFor
                \State $\mathbf{\hat{I}}, \boldsymbol{\hat{f}}$ $\gets$ $Aggregation(\mathbf{G}, \gamma)$ \\
            \Return $\mathbf{\hat{I}}, \boldsymbol{\hat{f}}$
		\EndProcedure
	\end{algorithmic}
\end{algorithm}

\textbf{(2) Graph connectivity-based sub-objects aggregation.}
This process is shown in lines 7-19 of Alg.~\ref{tab:algorithm}. We use the $s$ sub-objects obtained from the over-segmentation step as nodes to construct a graph and aggregate the sub-objects into complete objects by evaluating node connectivity.
This graph can be represented by an $s\times s$ adjacency matrix $\mathbf{G}$. $\mathbf{G} _{ij}$ represents the connectivity (edge) between the $i$-th and $j$-th sub-objects (nodes):
\begin{equation}
    \mathbf{G} _{ij} = \| \boldsymbol{\bar{f}}_i - \boldsymbol{\bar{f}}_j \|_2 \cdot \mathds{1}(i,j),
    \label{eq:graph}
\end{equation}
where we design two connectivity criteria: \textbf{1)} the L2 distance of the node features, a closer distance indicates that the two nodes are more likely to be merged; \textbf{2)} the spatial relationship between the nodes is represented by the indicator function $\mathds{1}(\cdot)$, which determines whether two sub-objects are adjacent or share voxels by voxelizing the 3D space with a resolution $r$, and returns 0 or 1.

After constructing the graph, we perform aggregation on the nodes with connectivity values $\mathbf{G} _{ij}$ in the range $(0, \gamma]$ using the standard graph connected component algorithm (such as breadth-first search), where $\gamma$ is set to $0.1$ in our implementation. Due to the indicator function, our graph is sparse and can be accelerated through the union-find method. Finally, by aggregating the feature-spatial similar sub-objects, we adaptively obtain several (\textit{e.g.}, $m$) independent and complete objects with $\boldsymbol{\hat{f}} \in \mathbb{R}^{m \times 6}$ representing their feature, $\mathbf{\hat{I}} \in \mathbb{Z}_{[0, m)}^n $ representing point label.

\section{Experiments*}
\footnotetext{$*$ Please refer to the \textcolor{red}{\textbf{supplementary materials}} for additional 3D/2D visualization results and comparisons with more methods.}
\subsection{Category-Agnostic 3D Instance Segmentation}

\textbf{\indent Settings.} \textbf{1) Task}: Category-agnostic 3D instance segmentation aims to segment independent 3D objects/instances in a given point cloud (Gaussians can naturally be treated as points) scene without considering their semantic categories, ensuring that each point ultimately belongs to only one instance.
\textbf{2) Dataset and Metrics}: Following OpenGaussian, we evaluate using the ScanNet~\cite{dai2017scannet} dataset with GT instance-level annotations. More specifically, we calculate the instance-level mIoU and mAcc metrics for the 10 scenes selected by OpenGaussian.
\label{baseline}\textbf{3) Baseline:} We mainly compare with Gaussian-based methods OpenGaussian~\cite{NEURIPS2024_21f7b745} and GaussianGrouping~\cite{gaussian_grouping} with RGB images and SAM masks as input. Additionally, we provide the performance of the depth-assisted back-projection method MaskClustering~\cite{yan2024maskclustering} as a reference. 

\begin{figure*}[t]
  \centering
  \includegraphics[width=0.9\linewidth]
    {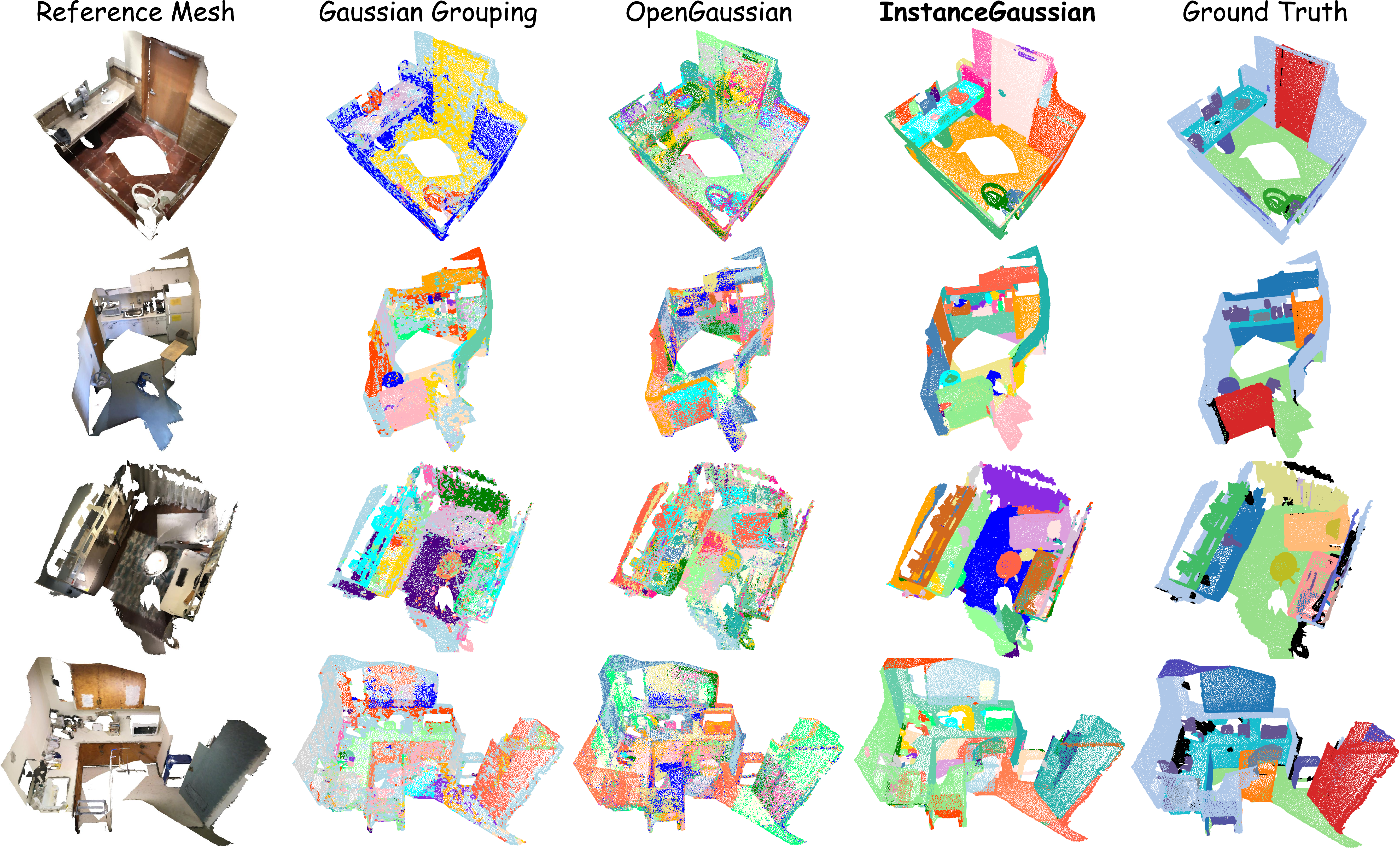} 
    \vspace{-3mm}
    \caption{Visualization comparison of category-agnostic 3D instance segmentation result. InstanceGaussian outperforms OpenGaussian and GaussainGrouping in accurately distinguishing different 3D objects.}
    \vspace{-3mm}
    \label{fig:instance}
\end{figure*}

\textbf{Results.} Quantitative results, as shown in Tab.~\ref{tab:instace_seg}, demonstrate that InstanceGaussian significantly outperforms Gaussian-based methods and achieves comparable results to MaskClustering which uses depth to assist. 
GaussianGrouping relies on a tracking-based model for cross-view object association, which is prone to tracking failures in complex scenes like ScanNet, characterized by significant viewpoint changes and occlusions, consequently leading to lower performance.
Although OpenGaussian achieves instance feature learning with cross-view 3D consistency by the smooth-contrast losses, avoiding explicit tracking-based association, its performance bottleneck lies in the brute-force, naive $k$-means clustering, lacking parameter adaptability.
Our method enables more discriminative feature learning through appearance-semantic joint representation and training, achieving more accurate and generalizable 3D instance segmentation through over-segmentation followed by connectivity-based adaptive aggregation.
The visualization results are shown in Fig.~\ref{fig:instance}, where the well-segmented instances faithfully reflect the performance advantages of InstanceGaussian.

\begin{table}[t]
\centering
\fontsize{9pt}{10.8pt}\selectfont
\resizebox{\columnwidth}{!}{
\begin{tabular}{c|c|cc}
    \toprule
    Type &Methods & mIoU $\uparrow$ & mAcc. $\uparrow$ \\ 
    \midrule
     \textcolor{gray}{Depth-aided} &\textcolor{gray}{MaskClustering~\cite{yan2024maskclustering}} & \textcolor{gray} {54.17} & \textcolor{gray} {80.95}  \\ 
    \midrule
     \multirow{3}{*}{\shortstack{Gaussian-\\based}}
     &GaussianGrouping~\cite{gaussian_grouping} & 22.55 & 30.54 \\ 
     &OpenGaussian~\cite{NEURIPS2024_21f7b745} & 27.32 & 52.44  \\ 
     &\textbf{InstanceGaussian (ours)} & \textbf{50.27} & \textbf{80.22}  \\ 
    \bottomrule
\end{tabular}
}
\vspace{-2mm}
\caption{Performance of category-agnostic 3D instance segmentation on the Scannet dataset. Accurate is measured by mAcc@0.25.}
\vspace{-4mm}
\label{tab:instace_seg}
\end{table}

\subsection{Open-Vocabulary Query Point Cloud Understanding}
\label{sec:exp_3d_semantic}
After 3D instantiation as mentioned in Sec.~\ref{sec:method_instantiation}, we adopt the 2D mask-3D instance association method proposed by OpenGaussian to associate each 3D instance with a 512-dim CLIP image feature. This allows us to support text-based understanding by extracting 512-dim CLIP text features and calculating cosine similarity.

\textbf{Settings.} \textbf{1) Task}: This task takes text as input and finds the matching point cloud by calculating the cosine similarity between text features and point features. In theory, it supports arbitrary open vocabulary input, but to compute quantitative metrics with the annotated GT point cloud, we use text that corresponds to the pre-defined categories as queries.
\textbf{2) Dataset and Metrics}: Following OpenGaussian, we use 19, 15, and 10 categories from ScanNet as text queries, assigning the closest text to the point cloud to compute the mIoU and mAcc for each category.
\textbf{3) Baseline:} We mainly compare with the recent Gaussian-based methods LangSplat, LEGaussians, and OpenGaussian. We also provide the performance of a SOTA method MaskClustering utilizing depth information to assist back projection as a comparison.
\begin{table}[t]
\centering
\resizebox{\columnwidth}{!}{
\begin{tabular}{c|c|cc|cc|cc}
\toprule
\multirow{2}{*}{Type} & \multirow{2}{*}{Methods} & \multicolumn{2}{c|}{19 classes} & \multicolumn{2}{c|}{15 classes} & \multicolumn{2}{c}{10 classes} \\
                      &                          & mIoU & mAcc. & mIoU & mAcc. & mIoU & mAcc. \\ \midrule
\textcolor{gray}{Depth} & \textcolor{gray}{MaskClustering~\cite{yan2024maskclustering}} & \textcolor{gray}{34.59} & \textcolor{gray}{54.32} & \textcolor{gray}{34.00} & \textcolor{gray}{45.07} & \textcolor{gray}{38.10} & \textcolor{gray}{56.74} \\ \midrule
\multirow{4}{*}{\begin{tabular}[c]{@{}c@{}}Gaussian-\\ based\end{tabular}} & LangSplat~\cite{qin2024langsplat} & 3.78 & 9.11 & 5.35 & 13.20 & 8.40 & 22.06 \\
                          & LEGaussians~\cite{shi2024language} & 3.84 & 10.87 & 9.01 & 22.22 & 12.82 & 28.62 \\
                          
                          &                        OpenGaussian~\cite{NEURIPS2024_21f7b745} & 24.73 & 41.54 & 30.13 & 48.25 & 38.29 & 55.19 \\
                          & \textbf{InstanceGaussian} & \textbf{40.66} & \textbf{54.01} & \textbf{42.51} & \textbf{59.15} & \textbf{47.94} & \textbf{64.01} \\ \bottomrule
\end{tabular}}
\vspace{-2mm}
\caption{Performance of open vocabulary semantic segmentation on the Scannet dataset. Accurate is measured by mAcc@0.25.}
\vspace{-5mm}
\label{tab:semantic_seg}
\end{table}

\begin{figure*}[t]
  \centering
  \includegraphics[width=0.9\linewidth]
    {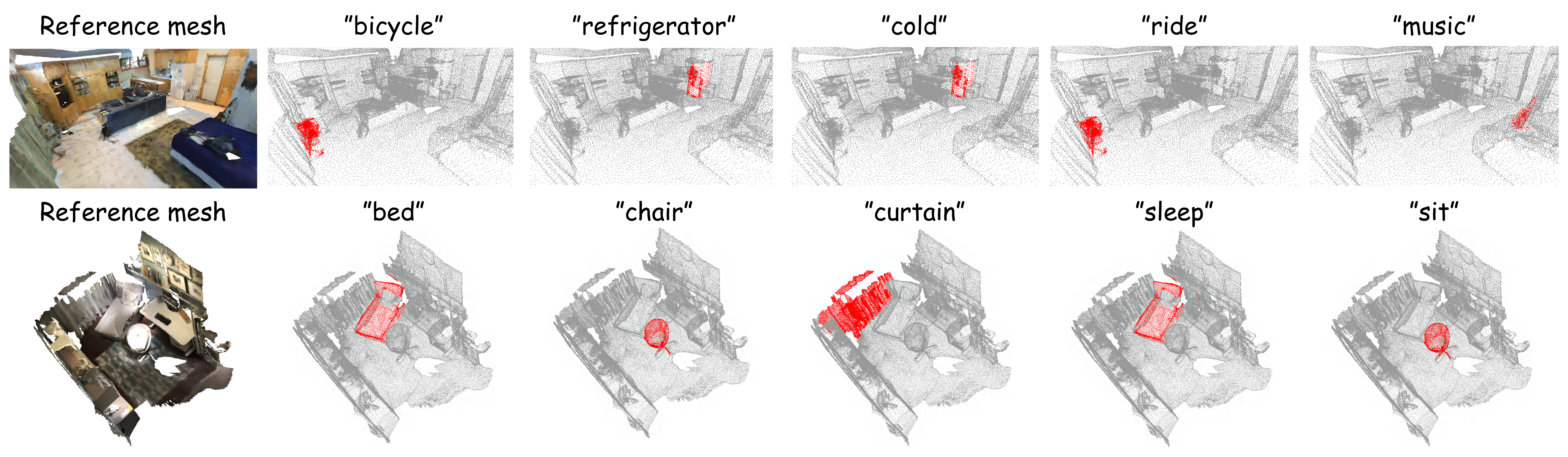} 
    \vspace{-3mm}
    \caption{Open-vocabulary query point cloud Understanding on ScanNet dataset. InstanceGaussian shows advanced text query capabilities.}
    \label{fig:openv} 
    \vspace{-4mm}
\end{figure*}

\textbf{Results.} 
The quantitative results are shown in Tab.~\ref{tab:semantic_seg} , where InstanceGaussian achieves SOTA performance under the settings of 19, 15, and 10 categories, significantly outperforming Gaussian-based methods.
The poor performance of LangSplat and LEGaussians stems from ambiguous Gaussian feature learning and detrimental language feature distillation.
The performance of OpenGaussian is limited by its coupled appearance-semantic representation and fixed cluster count.
GaussianGrouping is not included in the comparison because it does not support 3D-level text queries.
Surprisingly, our method even surpasses the depth-assisted MaskClustering, although it outperforms our method in the category-agnostic instance segmentation task. We analyze that the reason lies in its heavy reliance on high-quality SAM masks.
Fig.~\ref{fig:openv} shows the results of relevance point cloud retrieval using text queries, in which both common categories (such as ``bed'', ``curtain'') are used as queries, and the ability to understand open-vocabulary (such as ``ride'', ``music'') is also demonstrated.

\begin{table}[t]
\centering
\fontsize{9pt}{10.8pt}\selectfont
\begin{tabular}{c|cc}
\toprule
Methods & mIoU $\uparrow$ & mAcc. $\uparrow$ \\ 
\midrule
LangSplat~\cite{qin2024langsplat} & 9.66 & 12.41 \\ 
LEGaussians~\cite{shi2024language} & 16.21 & 23.82  \\ 
OpenGaussian~\cite{NEURIPS2024_21f7b745} & 38.36 & 51.34  \\ 
\textbf{InstanceGaussian (ours)} & \textbf{45.30} & \textbf{58.44}  \\ 
\bottomrule
\end{tabular}
\vspace{-2mm}
\caption{Performance of open vocabulary 3D object selection and rendering on Lerf dataset. Accurate is measured by mAcc@0.25.}
\vspace{-5mm}
\label{tab:Lerf}
\end{table}

\subsection{Open Vocabulary 3D Object Selection and \\Rendering}
\label{sec:exp_select_render}
\textbf{\indent Settings.} \textbf{1) Task}: Similar to Sec.~\ref{sec:exp_3d_semantic}, we first use open vocabulary text to select matching 3D Gaussians, and then render the 3D Gaussians into multi-view 2D images through the rasterization pipeline of 3DGS.
\textbf{2) Dataset and Metrics}: Following LangSplat and OpenGaussian, we use the LeRF~\cite{kerr2023lerf} dataset for evaluation, 
and after rendering the selected 3D objects into 2D images, we calculate the mIoU and mAcc between the rendered 2D objects and the GT 2D target objects.
\textbf{3) Baseline}: Only Gaussian-based methods possess 3D point-level perception and rendering capabilities, thus we only compare against Gaussian-based methods Langsplat, Legaussian, and OpenGaussian.

\begin{figure*}[!h]
  \centering
  \includegraphics[width=0.9\linewidth]
    {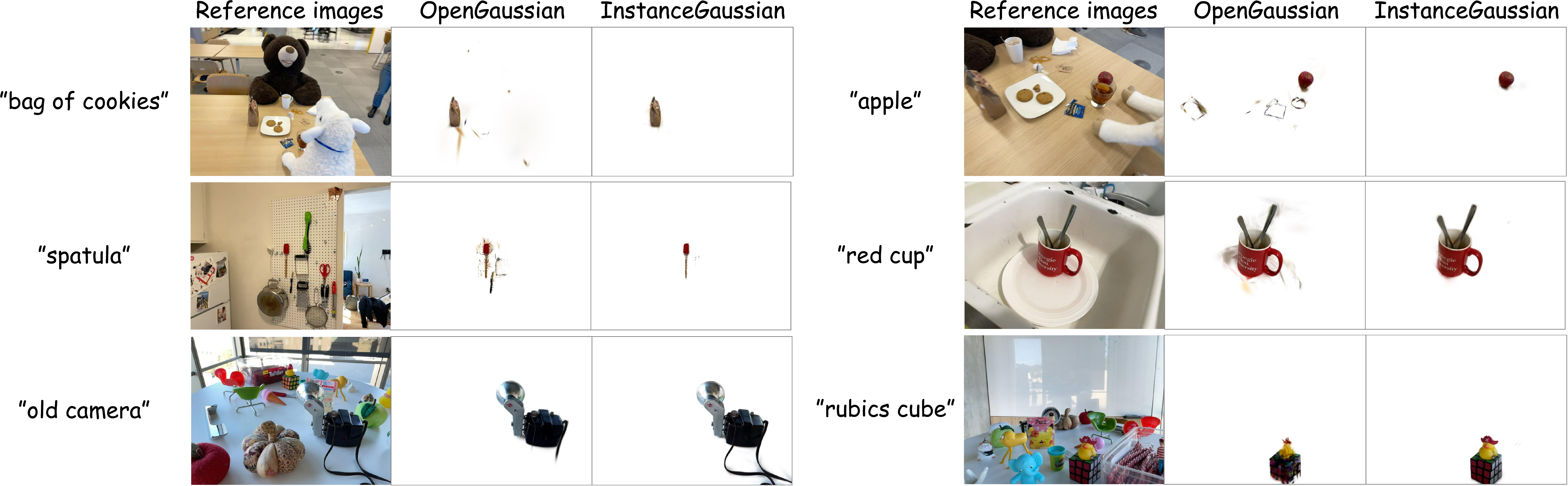} 
    \vspace{-2mm}
    \caption{Open-vocabulary 3D object selection and rendering on the LeRF dataset. InstanceGaussian outperforms OpenGaussian in accurately identifying the objects'  boundaries by text queries.}
    \label{fig:lerf}
    \vspace{-2mm}
\end{figure*}

\textbf{Result.} Quantitative results are displayed in the Tab.~\ref{tab:Lerf}, which demonstrates that even when rendering 3D Gaussians onto 2D images for evaluation.
As previously analyzed, LangSplat and LEGaussian, due to their poor 3D perception capabilities, fail to accurately select 3D Gaussians relevant to the text query, resulting in low-quality rendering results.
The performance advantage of our method stems from two aspects. First, the bottom-up adaptive instantiation ensures good discrimination between objects. Second, the joint representation and learning of appearance and semantics avoid the appearance-semantic inconsistency present in OpenGaussian.
The qualitative results in Fig.~\ref{fig:lerf} more intuitively demonstrate this point. Our method renders objects with clearer boundaries, unlike OpenGaussian, which due to its appearance-semantic ambiguity, requires some Gaussians to represent multiple objects/regions, especially evident in the boundary areas.

\begin{figure*}[!h]
  \centering
  \includegraphics[width=0.9\linewidth]
    {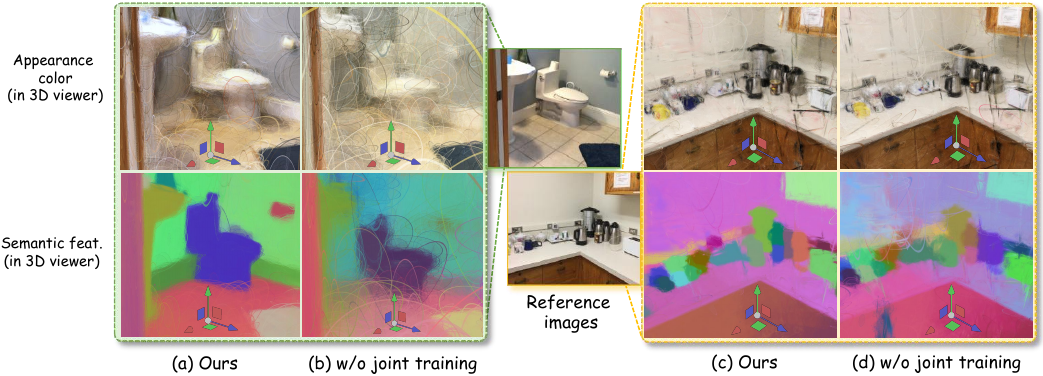}
    \vspace{-2mm}
    \caption{Visualization results of the appearance color and instance features of the Gaussians ellipsoid in the 3D viewer demonstrate that our method achieves superior appearance-semantic consistency. (a)(c): Results of progressive joint training. (b)(d): Without joint training.}
    \label{fig:vis_joint_ablation}
    \vspace{-3mm}
\end{figure*}

\subsection{Ablation Study}

\subsubsection{Gaussian Representation and Training Strategy}

We conduct the ablation studies as shown in Tab.~\ref{tab:ab_joint_train} to demonstrate the effectiveness of our Semantic-Scaffold-GS joint representation and progressive joint training strategy. The experiments are evaluated using the semantic and instance segmentation tasks on the ScanNet dataset.
The joint representation achieves a balance between appearance and semantics, allowing the same anchor point to express different appearances while sharing the same semantics.
The progressive joint training avoids inconsistencies in appearance and semantics, enhances the distinguishability between objects, and thus leads to improved instance segmentation and semantic understanding performance.

Fig.~\ref{fig:vis_joint_ablation} more intuitively demonstrates our avoidance of inconsistency. The first and second rows respectively visualize the appearance color and instance features of the 3D Gaussian ellipsoid in the 3D viewer, where the ellipse represents the boundary of the Gaussian.
Fig.~\ref{fig:vis_joint_ablation} (a)(c) shows the results of our joint training, which achieve clearer and more distinctive expressions in both appearance and semantics compared to Fig.~\ref{fig:vis_joint_ablation} (b)(d). 
This improvement not only brings benefits to the understanding of the 3D level (\textit{i.e.}, the semantic level, as proven by our quantitative ablation studies), but also enhances the quality of object rendering (\textit{i.e.}, the appearance level, as shown in Sec.~\ref{sec:exp_select_render}).
\begin{table}[t]
\small
\centering
\resizebox{\columnwidth}{!}{
\begin{tabular}{c|cc|cc}
\toprule
Case & Jotin-repres. & Joint-train & Sematic Seg. & Instance Seg. \\ \midrule
\#1  & \ding{51}     &             & 30.71        & 47.40          \\
\#2  &               & \ding{51}   & 33.15        & 49.57         \\
\#3  & \ding{51}     & \ding{51}   & 40.66        & 50.27         \\ \bottomrule
\end{tabular}
}
\vspace{-2mm}
\caption{Performance of instance and semantic segmentation with different representation and training strategies on Scannet. Evaluated by mIoU. }
\vspace{-4mm}
\label{tab:ab_joint_train}
\end{table}

\begin{table}[t]
\centering
\fontsize{9pt}{10.8pt}\selectfont
\resizebox{\columnwidth}{!}{
        \begin{tabular}{c|cc|cc}
        \toprule
        Case
                              & feature                & voxelize                  & Semantic Seg.                          &        Instance Seg.                   \\ \midrule
        \#1                   & \multicolumn{1}{l}{} & \ding{51}             & 21.75                     & 27.50                     \\
        \#2                   & \ding{51}            & \multicolumn{1}{l|}{} & 28.98                     & 43.41                     \\
        \#3                   & \ding{51}            & \ding{51}             & 40.66                     & 50.27                     \\ \bottomrule
        \end{tabular}}
    \vspace{-2mm}
    \caption{Performance of instance and semantic segmentation with different connectivity condition strategies. Evaluated by mIoU.}
    \vspace{-5mm}
    \label{tab:connectivity_condition}
\end{table}
\subsubsection{Aggregation Condition}

In the process of aggregating over-segmented sub-objects based on a connectivity graph, we considered two connectivity conditions: the L2 distance of instance features, and the proximity judgment after voxelization.

We conduct an ablation study on its configuration as shown in Tab.~\ref{tab:connectivity_condition}. In case \#1, we use the color of the Gaussians as a feature and consider their positions for aggregation, which results in poor performance. In case \#2, we only consider instance features and achieve a significant performance improvement, demonstrating the effectiveness of our feature learning. In case \#3, we take both conditions into account and achieve the best performance, which is an intuitive idea that in aggregation, we need to consider both semantic similarity and spatial proximity.

\section{Conclusion and Limitation}
\vspace{-2mm}
In this paper, we introduce an appearance-semantic compact Gaussian representation to address the imbalance between appearance and instance features using Gaussians. On one hand, this representation enhances the learning of appearance; on the other hand, sharing features among neighboring points significantly boosts the efficiency of semantic learning. The bottom-up category-agnostic instantiation adaptively determines the number of categories, cleverly avoiding the issue of over-segmentation or under-segmentation. In category-agnostic segmentation tasks, we significantly outperform previous models, achieving state-of-the-art (SOTA) performance in open vocabulary understanding for 3D point-level tasks.

Despite achieving novel performance on multiple tasks, our method still has the following limitations: \textbf{1)} Longer training time for slower rendering, since the appearance attributes need to be calculated through MLP. \textbf{2)} Segmentation accuracy is still affected when most of the images in the scene are incorrectly segmented by SAM. Failure case analyses can be found in the \textcolor{red}{\textbf{supplementary materials}}. We hope those drawbacks can be addressed in future works.

\newpage
\appendix




\newpage

\section{Implementation Details}

\subsection{Training Strategy.}

For ScanNet~\cite{dai2017scannet} dataset, we freeze the point cloud coordinates and disable 3DGS~\cite{kerbl20233d} densification.
For the LeRF~\cite{kerr2023lerf} dataset, we optimize the point cloud coordinates and enable 3DGS densification. We stop 3DGS densification in 10k steps.

\subsection{Training Time}
We train each scene on a single 24G 3090 GPU (with actual memory usage around 5 to 10 GB). For the LeRF dataset, each scene takes around 200 images and trains for approximately 70 minutes. For the ScanNet dataset, each scene takes around 100-300 images and trains for approximately 30 minutes. Tab.~\ref{tab:training time} shows the comparison with the baseline method in training time.  We selected level 3 to extract the SAM~\cite{sam} mask.

\begin{table}[h]
    \small
    \centering
    {
    \begin{tabular}{ccccc}
    \hline
    Method & Langslpat & LEGaussian & OpenGaussian & Ours                                \\ \hline
     Time (s)& 1153    &  1011  & 1378    &  1847            \\ \hline
    \end{tabular}}
    \caption{Comparation of training time in Scannet.}
    \label{tab:training time}
\end{table}

\subsection{ScanNet Dataset Setting}
We randomly selected 10 scenes from ScanNet for evaluation,
specifically: scene0000\_00, scene0062\_00, scene0070\_00, scene0097\_00, scene0140\_00, scene0200\_00, scene0347\_00, scene0400\_00, scene0590\_00, scene0645\_00. The 19 categories (defined by ScanNet) used for text query are respectively: wall, floor, cabinet,
bed, chair, sofa, table, door, window, bookshelf, picture, counter, desk, curtain, refrigerator, shower curtain, toilet, sink, bathtub; 15 categories are without picture, refrigerator, showercurtain, bathtub; 10 categories are further without cabinet, counter, desk, curtain, sink. We downsampled the training images by a factor of 2 and selected SAM level 3 to extract supervision signals.

\noindent For the fixed point clouds in the ScanNet dataset, suboptimal processing often leads to degraded visual quality. To address this, we identified well-optimized point clouds based on their contributions during rendering and utilized them to train a lightweight MLP. The MLP takes position and color as inputs and predicts segmentation labels as outputs. Subsequently, the trained MLP is employed to infer segmentation results for the point clouds. This approach yields smoother segmentation outcomes, significantly enhancing visual quality.

\section{Comparison with more Methods}
We conducted comparisons with additional baselines in Tab.~\ref{tab:Comparison}, demonstrating the superior performance of our method.
\noindent
\begin{table}
\centering
\small
{
\begin{tabular}{ccccc}
\hline

                         & \multicolumn{2}{c}{Semantic seg.} & \multicolumn{2}{c}{Instance seg.} \\
\multirow{-2}{*}{Method} & mIoU                        & mAcc.                       & mIoU                        & mAcc.                       \\ \hline
GAGA\cite{lyu2024gaga}                                             & \multicolumn{2}{c}{\textit{-unsupported-}}                  & 16.76                            &31.49                              \\
SAGA\cite{cen2023saga}                                             &         9.44               & 16.23                       &   34.16                      &        70.14                \\
\textbf{Ours}                                    & \textbf{40.66}              & \textbf{54.01}              & \textbf{50.27}              & \textbf{80.22}             \\ \hline
\end{tabular}}
    \captionof{table}{Comparison with more methods.}
    \label{tab:Comparison}
\end{table}%

\section{Ablation Study}
\subsection{More Detailed Ablation Metrics}
We conducted a detailed ablation study with mA50 and mA25 metric in Tab.~\ref{tab:Additional_ablation} and Tab.~\ref{tab:Additional_ablation2}. We chose joint representation and joint training as the best strategy for the segmentation topic, as mIoU is generally more reasonable than mAcc. 
\begin{table}[h]
\centering
\small
{
\begin{tabular}{ccccccc}
\hline
\multicolumn{2}{c}{Joint} & \multicolumn{2}{c}{Semantic seg} & \multicolumn{3}{c}{Instance Seg} \\
Rep.                      & Train.                     & mIoU                      & mAcc                     & mIoU             & mA50             & mA25\\ \hline
\ding{51}                                                       &                                                       & 30.71                       & 44.22                      & 47.4              & 44.51             & \textbf{84.42}            \\
& \ding{51}                                                  & 33.15                       & 44.45                      & 49.57             & \textbf{52.82}             & 82.17            \\
\ding{51}                                                       & \ding{51}                                                     & \textbf{40.66}                       & \textbf{54.01}                      & \textbf{50.27}             & 52.57             & 80.22            \\ \hline
\end{tabular}}
\caption{Detailed ablation of joint representation and training.}
    \label{tab:Additional_ablation}
\end{table}

\begin{table}[h]
\small
\centering
{
\begin{tabular}{ccccccc}
\hline
\multicolumn{2}{c}{Condition} & \multicolumn{2}{c}{Semantic seg} & \multicolumn{3}{c}{Instance Seg} \\
Feat.                      & Vox.                     & mIoU                      & mAcc                     & mIoU             & mA50             & mA25             \\ \hline
                           & \ding{51}                        & 21.75                     & 33.39                    & 27.50            & 16.16           & 49.33           \\
\ding{51}                          &                          & 28.98                     & 40.58                    & 43.41            & 40.63           & 72.50           \\
\ding{51}                          & \ding{51}                        & \textbf{40.66}            & \textbf{54.01}           & \textbf{50.27}   & \textbf{52.57}  & \textbf{80.22}  \\ \hline
\end{tabular}}
\caption{Detailed ablation of aggregation condition.}
    \label{tab:Additional_ablation2}
\end{table}

\subsection{Hyperparameters}
We conducted ablation experiments of hyperparameters in Tab.~\ref{tab:hyperparameters}, demonstrating the robustness of our instantiated algorithm. The hyperparameters reported in our paper were not carefully tuned, and we discovered even better performance when adjusting them.
\begin{table}[h]
\small
\centering{
\resizebox{\columnwidth}{!}{
\begin{tabular}{cccc}
\hline
                                  & sample number & voxel size     & connectivity thredshold \\
\multirow{-2}{*}{hyperparameters} & 200$\sim$1000 & 0.05$\sim$0.5 & 0.06$\sim$0.18          \\  \hline
semantic seg. mIoU  & 37.87$\pm$2.79     & 38.96$\pm$1.70   & 38.48$\pm$2.17             \\
instance seg. mIoU & 49.60$\pm$1.31     & 49.56$\pm$0.71   & 49.90$\pm$1.29             \\ \hline
\end{tabular}}}
\caption{Abliation of hyperparameters.}
    \label{tab:hyperparameters}
\end{table}
\subsection{Randomness and Runtime of FPS}
We randomly sampled different starting points for FPS in Tab.~\ref{tab:random}, and the results demonstrated the robustness of our method.  We also evaluated runtime across different sample number ($K$) in Tab.~\ref{tab:K}. While the merging algorithm has a complexity of O($K^2$), $K$ shows limited impact on overall runtime in practice (up to 3.43\%). 

\begin{table}[h]
    \small
    \centering{
    \begin{tabular}{ccc}
    \hline
    Sample times & Semantic Seg. mIoU                 & Instance Seg. mIoU                 \\ \hline 5& 
              39.44$\pm$1.38            & 48.57$\pm$0.98                  \\ \hline
    \end{tabular}}
    \caption{Ablation experiment of randomness in FPS.}
    \vspace{-3mm}
    \label{tab:random}
\end{table}

\begin{table}[h]
\small
\centering{
\resizebox{\columnwidth}{!}{
\begin{tabular}{cccc}
\hline
Variation           & { Merging time(s)} & { Total time(s)} & { Percentage} \\ \hline
 200$\sim$1000  & 5$\sim$63                                         & 1772$\sim$1847                                      & 0.28\%$\sim$3.43\%                                    \\ \hline
\end{tabular}}}
\caption{Ablation experiment of different $K$.}
    \label{tab:K}
\end{table}

\subsection{Aggregation Algorithm}
We compare our aggregation algorithm against clustering features by HDBSCAN\cite{campello2013density} in Tab.~\ref{tab:Ablation_HDBSCAN}, demonstrating our superiority.
\begin{table}[h]
    \small
    \centering{
    \begin{tabular}{ccc}
    \hline
    Method &Semantic Seg. mIoU                 &Instance Seg. mIoU                 \\ \hline
          HDBSCAN  &     31.84                &   44.76                 \\ 
    Ours  & \textbf{40.66}  & \textbf{50.27} \\ \hline
    \end{tabular}}
    \captionof{table}{Ablation of aggregation algorithm.}
    \vspace{-3mm}
    \label{tab:Ablation_HDBSCAN}
\end{table}
\section{More Visual Results}
\begin{figure*}[t]
  \centering
  \includegraphics[width=0.9\linewidth]
    {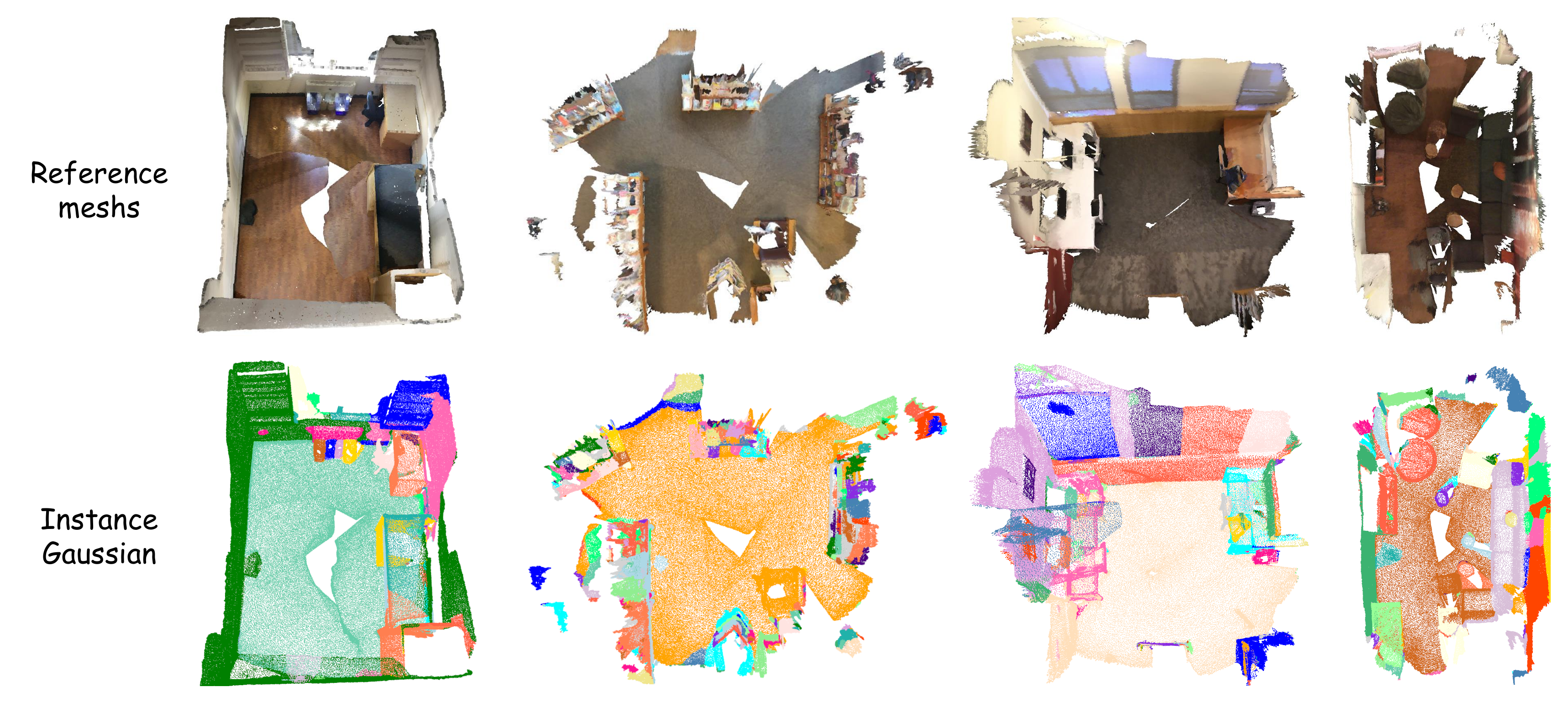}
    \caption{Top row: Reference mesh of scenes. Bottom row: The visualization result of category-agnostic 3D instance segmentation in ScanNet dataset.}
    \label{fig:supp_instance}
\end{figure*}

\begin{figure*}[t]
  \centering
  \includegraphics[width=0.9\linewidth]
    {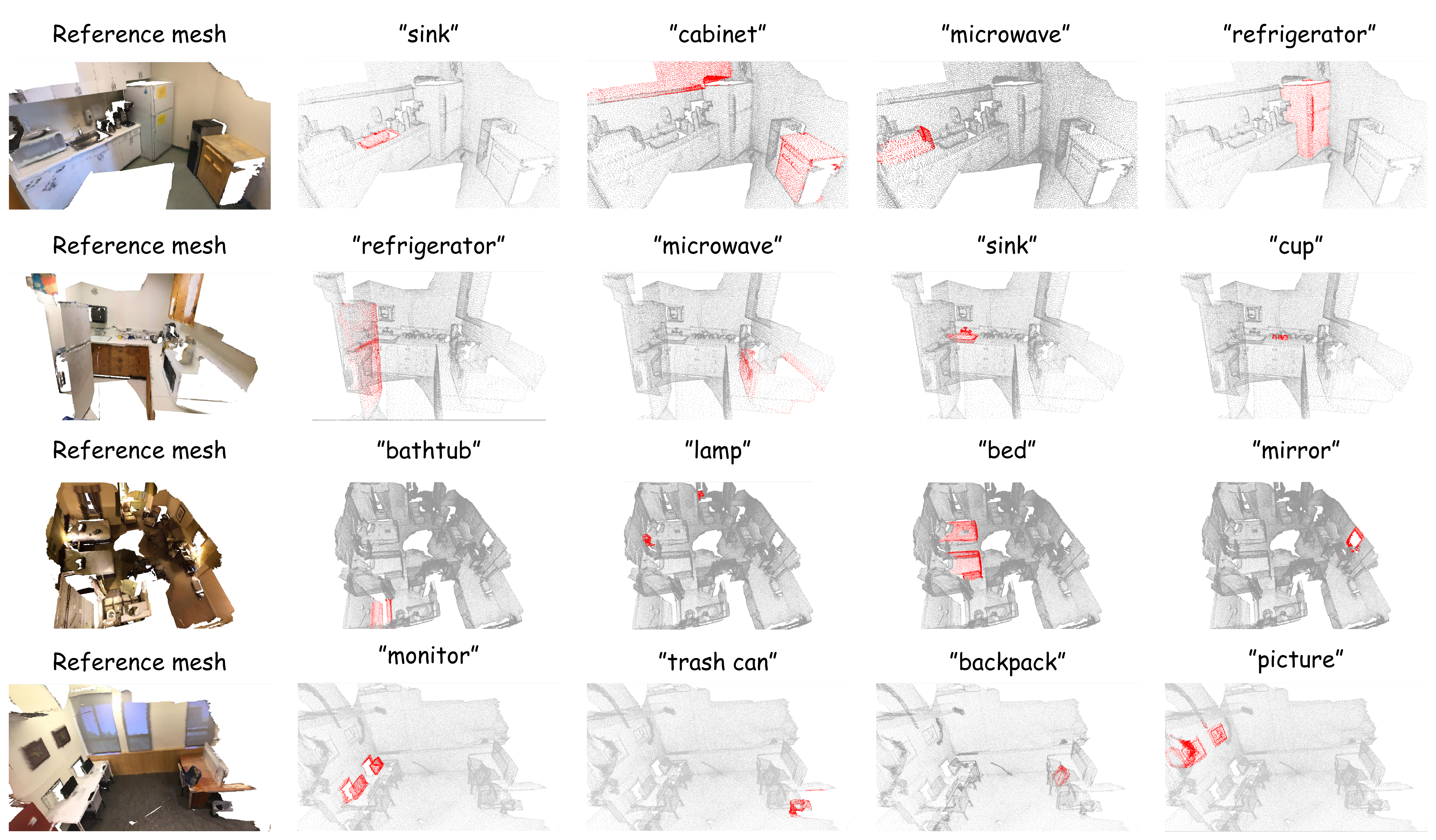}
    \caption{The visualization result of open-vocabulary query point cloud understanding in ScanNet~\cite{dai2017scannet} dataset.}
    \label{fig:supp_opev}
\end{figure*}

\begin{figure*}[t]
  \centering
  \includegraphics[width=0.9\linewidth]
    {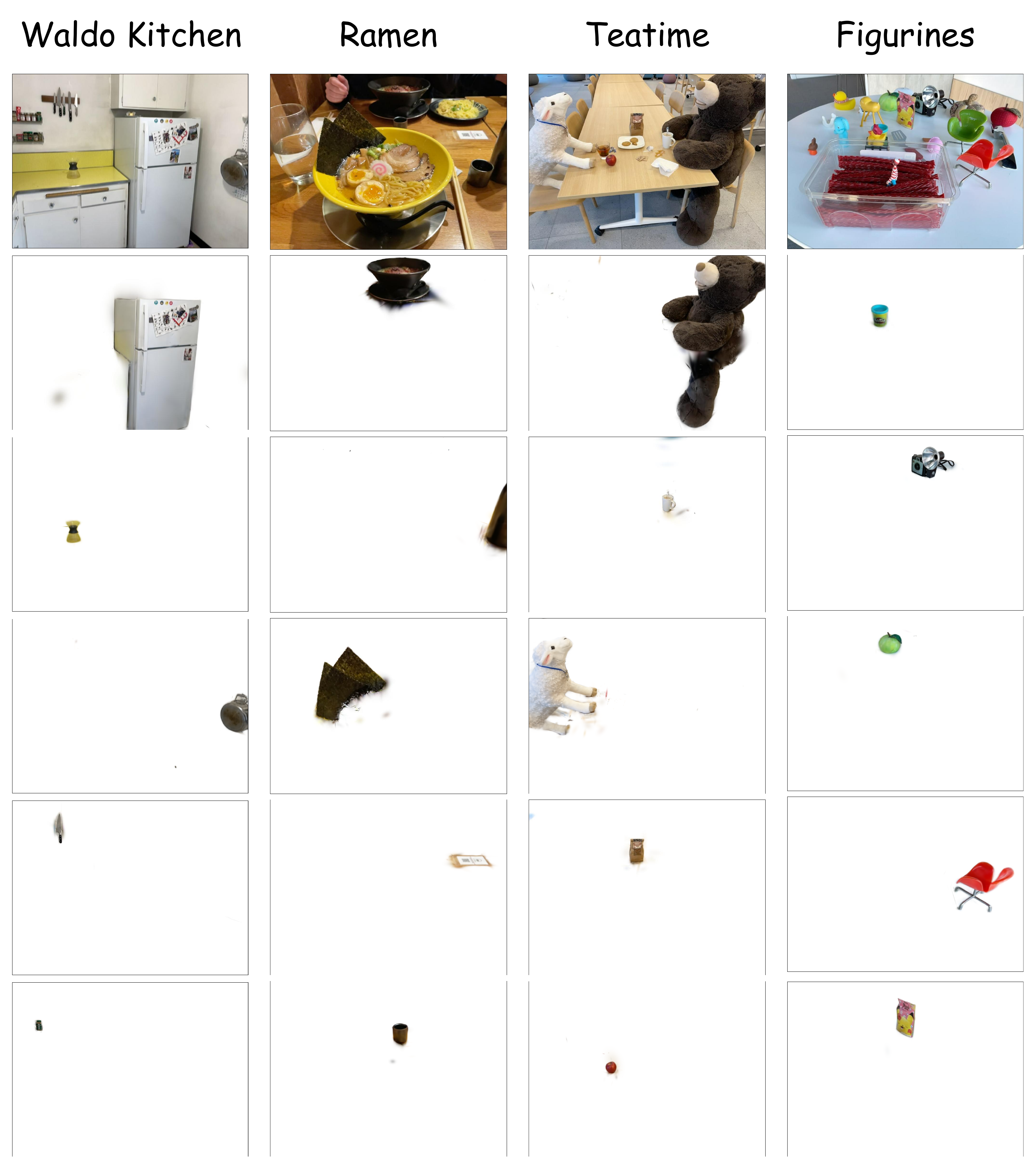}
    \caption{The 2D visualization result of 3D instance segmentation in LeRF dataset.}
    \label{fig:supp_lerf}
\end{figure*}

\begin{figure*}[t]
  \centering
  \includegraphics[width=0.9\linewidth]
    {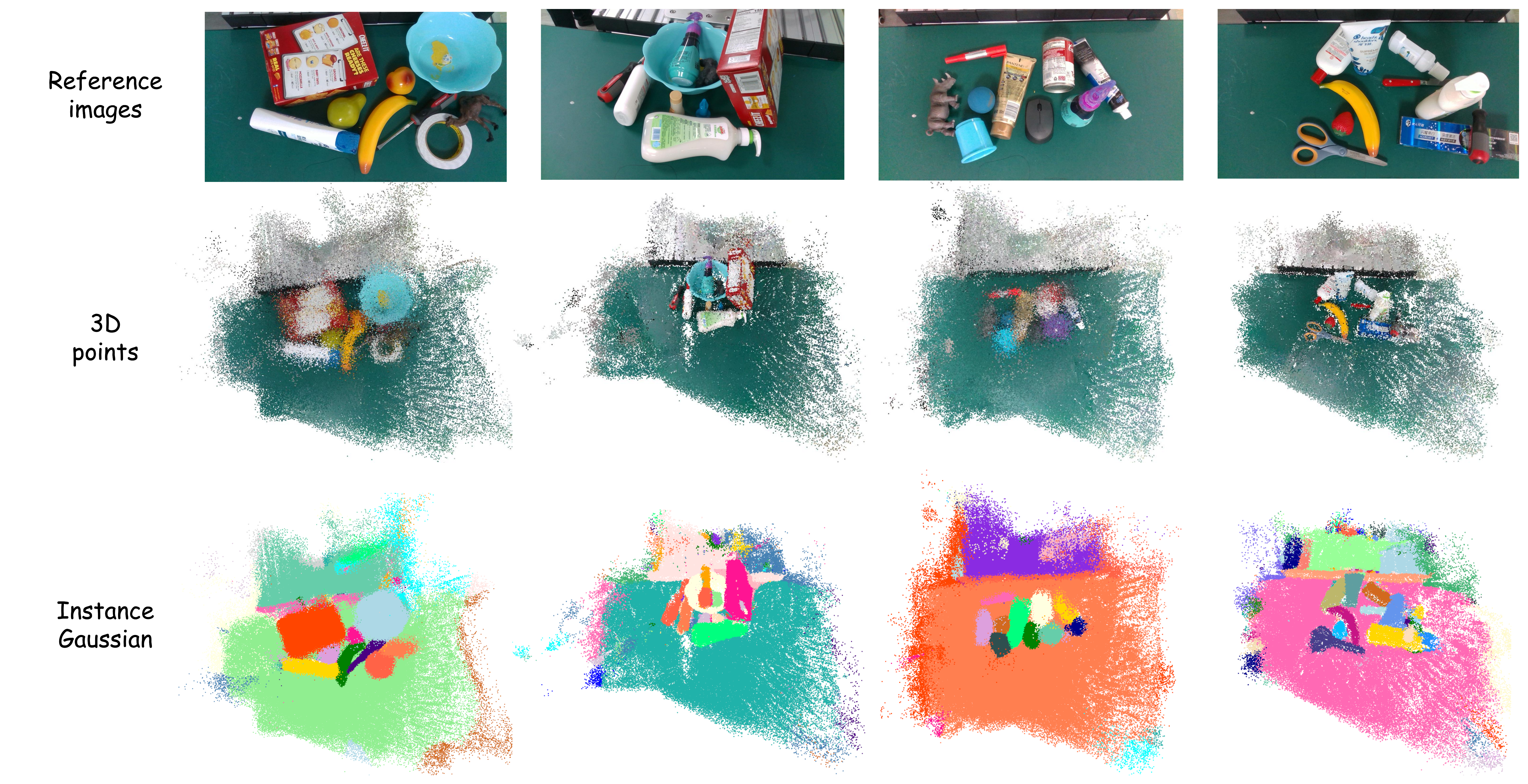}
    \caption{ Top: Reference image of scenes. Middle: Constructed 3D Gaussians/points. Bottom: The visualization result of category-agnostic 3D instance segmentation in GraspNet dataset.}
    \label{fig:grasp}
\end{figure*}

To demonstrate the effectiveness of our approach, we conducted additional experiments on ScanNet\cite{dai2017scannet} to prove our performance on category-agnostic 3D instance segmentation (Fig.~\ref{fig:supp_instance}) and open-vocabulary query point cloud understanding (Fig.~\ref{fig:supp_opev}). Additionally, we provide more 2D instance segmentation results on LeRF\cite{kerr2023lerf} (Fig.~\ref{fig:supp_lerf}).  We also performed experiments on GraspNet~\cite{fang2020graspnet} dataset (Fig.~\ref{fig:grasp}), and the results indicate the generalization capabilities of our method.

\begin{figure*}[t]
  \centering
  \includegraphics[width=0.75\linewidth]
    {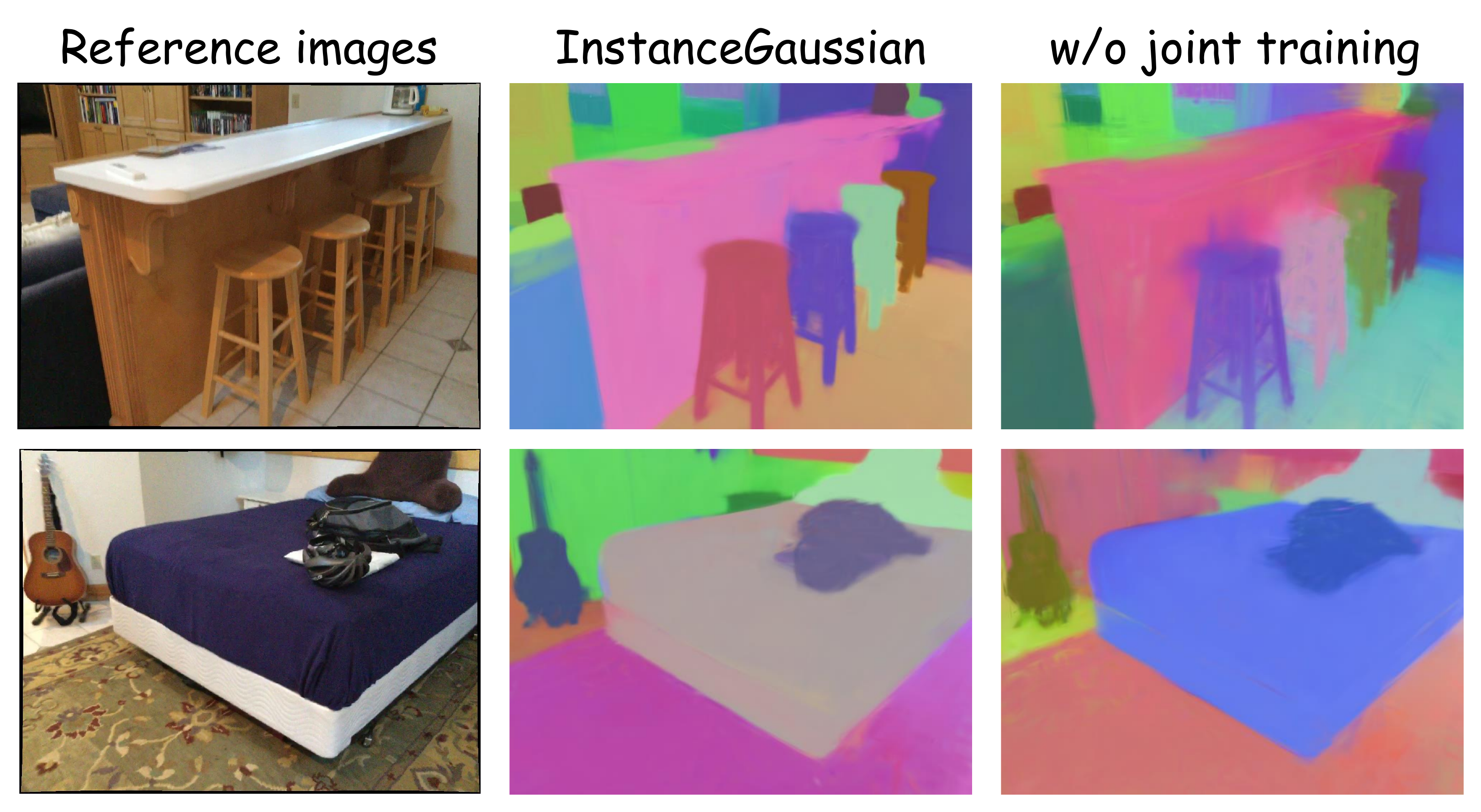}
    \caption{The feature map of InstanceGaussian. Left: Reference images of the scene. Middle: The visualization result of feature maps of InstanceGaussian. Right: The visualization result of feature maps without joint training.}
    \label{fig:featmap}
\end{figure*}

\noindent Additionally, we visualized our feature maps (Fig.~\ref{fig:featmap}) to validate the effectiveness of appearance-semantic joint Gaussian representation.

\section{Analysis of Failure Cases}
We analyze two main issues affecting the model's performance.

\begin{figure*}[t]
  \centering
  \includegraphics[width=0.9\linewidth]
    {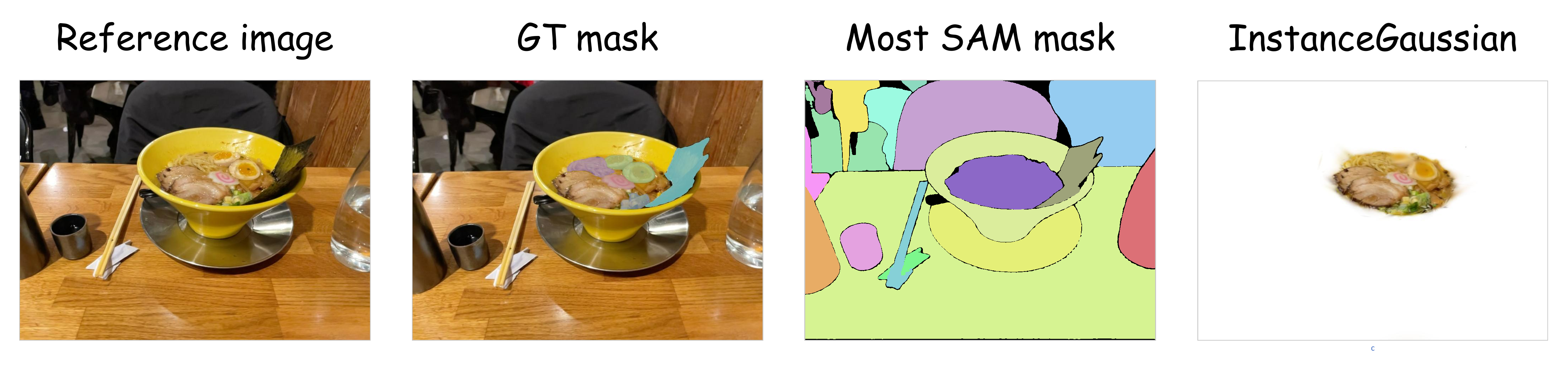}
    \caption{The failure case of ramen in LeRF dataset. Frequent mask failures will undermine the ability to distinguish different food in the bowl.}
    \label{fig:failure}
\end{figure*}

\textbf{Incorrect SAM mask}. Our method demonstrates robustness against sparse segmentation errors when SAM masks are predominantly accurate. However, frequent mask failures prevent effective learning of region-specific object features, ultimately compromising segmentation accuracy (Fig.~\ref{fig:failure}).

\begin{figure*}[t]
  \centering
  \includegraphics[width=0.9\linewidth]
    {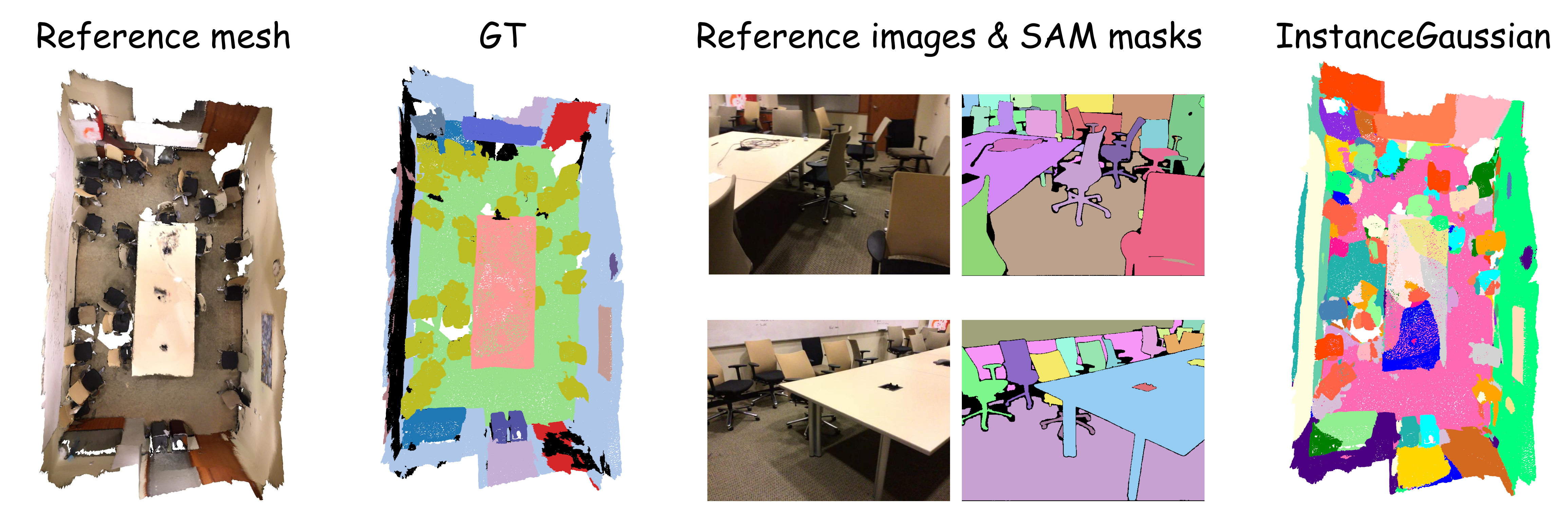}
    \caption{The failure case of aggregate the large meeting table in scene0140\_00. Due to the table's large size and the lack of fully captured views, our method struggles to learn consistent features and achieve accurate aggregation, even when the SAM mask is correct.}
    \label{fig:failure2}
\end{figure*}

\textbf{Failure in learning consistent feature for large object}. Our method demonstrates superior performance in aggregating small objects (e.g., doors, chairs, televisions) but encounters challenges with large-scale objects (e.g., floors, meeting tables). Due to the limited coverage of individual photographs, large objects are often only partially captured, with few complete observations across views. Under such conditions, $\mathcal{L}_s$ (Eq. 2) fails to learn consistent features for complete objects and $\mathcal{L}_c$ (Eq. 3) amplifies the divergence between parts (Fig.~\ref{fig:failure2}). In contrast, small objects benefit from more complete observations, enabling $\mathcal{L}_s$ (Eq. 2) to learn coherent features.

{   
    \clearpage
    \clearpage
    \small
    \bibliography{main}
}

\end{document}